\documentclass{article}
\usepackage{ijcai25}

\usepackage{times}
\usepackage{soul}
\usepackage{url}
\usepackage[hidelinks]{hyperref}
\usepackage[utf8]{inputenc}
\usepackage[small]{caption}
\usepackage{graphicx}
\usepackage{amsmath}
\usepackage{amsthm}
\usepackage{booktabs}
\usepackage{algorithm}
\usepackage{algorithmic}
\usepackage[switch]{lineno}
\usepackage{multirow}
\usepackage{boxedminipage}
\usepackage{arydshln} % 在表格中绘制局部的水平虚线

\title{Redefining Simplicity: Benchmarking Large Language Models from Lexical to Document Simplification
}

\author{
 Jipeng Qiang$^1$
\and
Minjiang Huang$^1$\and
Yi Zhu$^1$ \and 
Yunhao Yuan$^1$\and 
Chaowei Zhang$^1$\and 
Kui Yu$^2$\\
\affiliations
$^1$School of Information and Engineering, Yangzhou University, China\\
$^2$School of Computer Science and Information Technology, Hefei University of Technology, China\\
\emails
\{jpqiang, zhuyi, yhyuan, cwzhang\}@yzu.edu.cn, mz120231035@stu.yzu.edu.cn,
yukui@hfut.edu.cn
}

\begin{document}

\maketitle

\begin{abstract}
Text simplification (TS) refers to the process of reducing the complexity of a text while retaining its original meaning and key information. Existing work only shows that large language models (LLMs) have outperformed supervised non-LLM-based methods on sentence simplification. This study offers the first comprehensive analysis of LLM performance across four TS tasks: lexical, syntactic, sentence, and document simplification. We compare lightweight, closed-source and open-source LLMs against traditional non-LLM methods using automatic metrics and human evaluations. Our experiments reveal that LLMs not only outperform non-LLM approaches in all four tasks but also often generate outputs that exceed the quality of existing human-annotated references. Finally, we present some future directions of TS in the era of LLMs. 
\end{abstract}

\section{Introduction}

Text simplification (TS) aims to reduce the complexity of text while preserving its original meaning, thereby enhancing accessibility for low-literacy individuals, non-native speakers, and people with cognitive impairments \cite{xu-etal-2015-problems}. Over the past few decades, TS has gained significant attention, not only in English texts but also in dozens of other languages, including French \cite{koptient2020fine,ormaechea2024automatic}, Chinese \cite{qiang2023chinese,chong2024mcts}, Spanish \cite{bott2012text,saggion2015making}, and Japanese \cite{hayakawa2022jades,nagai2024document}. Furthermore, TS research spans various domains; beyond the common application to news articles and encyclopedias, it has also been utilized in specialized fields such as medical \cite{devaraj2021paragraph}, legal \cite{garimella2022text}, and patent \cite{kang2018text} texts, addressing domain-specific challenges.

The field of TS can be broadly categorized into four main research directions: (1) lexical simplification (LexS) \cite{paetzold2016unsupervised,qiang2021lsbert} focuses on replacing complex or uncommon words with simpler or more familiar alternatives, ensuring accessibility without losing meaning; (2) syntactic simplification (SynS) \cite{niklaus2023discourse,gao2021abcd} targets grammatical complexity by simplifying syntax, reducing intricate constructs while maintaining grammatical correctness and coherence; (3) sentence simplification (SenS) \cite{xu2016optimizing,martin-etal-2022-muss} aims to rephrase complex sentences into simpler sentences while retaining original meaning by changing its structure and content; (4) document simplification (DocS) \cite{sun-etal-2021-document,fang-etal-2025-collaborative} deals with simplifying an entire text while preserving its overall structure and logical flow, ensuring consistency and coherence across paragraphs and chapters.

Before the emergence of large language models (LLMs), text generation tasks, such as machine translation \cite{deng2024deep}, document summarization \cite{saxena2024improving}, and TS \cite{al2021automated}, relied more on fine-tuning sequence-to-sequence models using supervised data, achieving significant success . With the advent of LLMs, these models have revolutionized many text generation tasks, eliminating the need for additional fine-tuning. A single LLM has already achieved comparable or even superior performance to non-LLM approaches in many text generation tasks \cite{xu2024contrastive}. Numerous papers have conducted comprehensive analyses of LLM performance in machine translation \cite{zhang2023prompting} and document summarization \cite{zhang2024benchmarking}. However, existing research only shows that Previous studies have only demonstrated that large language models (LLMs) surpass supervised non-LLM-based approaches in sentence simplification, with no papers providing comparative analyses of LLM performance across all TS tasks.

The field of LLMs has garnered widespread attention from both industry and academia, with the performance of LLMs, whether open-source or closed-source, improving every few months. Inevitably, these advancements also enhance their capabilities in TS tasks. This raises two potential questions: (1) How well do LLMs perform across various TS tasks? (2) As LLMs continue to improve, is there still a need for research in TS? If so, what are the suitable research directions? This paper will answer these questions.

(1) To the best of our knowledge, this paper is the first to comprehensively analyze and compare the performance of LLMs across four tasks (LexS, SynS, SenS, and DocS). Lightweight (Gemma2-2B), closed-source (GPT-4o), open-source (Llama3.1-70B) LLMs are selected as baseline methods. In addition to commonly used automatic evaluation metrics, human evaluation was also conducted. The experimental results reveal the following findings: 1) Large-scale LLMs outperform non-LLM methods and human-annotated results across all four tasks. 2) Lightweight LLM excels in SenS and SynS tasks.
3) Existing automated evaluation datasets no longer meet the needs of LLMs, as the results generated by LLMs surpass the human-annotated references in some tasks. The prompts used for the LLM  and the experimental results are available \footnote{https://github.com/9624219/A-Comprehensive-Study-of-Large-Language-Models-in-Text-Simplification}. 

(2) As LLM performance continues to improve, the performance of LLMs in TSwill only continue to improve. Future work needs to focus on more complex text simplification tasks. We have proposed four possible research directions, including multi-level TS; personalized simplification tailored to individual users’ preferences and reading abilities; lightweight LLM training methods for more efficient model deployment; and automatic evaluation metrics without human annotations. 

\section{Text Simplification}

We conduct experimental analyses on four mainstream tasks of TS: LexS, SenS, SynS, and DocS. The experimental results present the performance of the best non-LLM-based TS approaches as well as methods utilizing large language models proposed in the latest research. Here, we employ three different types of language modelings as the baselines: (1) lightweight LLM, "Gemma2(2B)" via Hugging Face \footnote{https://huggingface.co/google/Gemma2-2b}; (2) open-source LLM, “Llama3.1(70B)” via Ollama package\footnote{https://ollama.com/}, and (3) closed-source LLM, “GPT-4o-2024-11-20” via OpenAI API\footnote{https://platform.openai.com/docs/models}. 

\subsection{Lexical Simplification (LexS)}

\textbf{(1) Baselines}

LexS task involves first identifying complex words and then finding the best substitutes for those complex words \cite{paetzold2016unsupervised}. LexS methods generally require the following three steps: complex word identification, substitute generation, and substitute ranking. A suitable substitute needs to align with the contextual information of the complex word and makes the sentence simpler and easier to understand. We choose the following methods for comparison.

\textbf{LSBERT.} LSBERT \cite{qiang2021lsbert} is one BERT-based LexS method, which involves masking the complex word and predicting potential substitutions based on the context.

\textbf{MANTI.} MANTI \cite{li2022mantis} is also BERT-based LexS method, which is one best method from (TSAR-2022) shared task\cite{saggion2023findings}.

\textbf{LSPG.} LSPG \cite{LiuQLYZH23} is a multilingual LexS method via paraphrase generation for generating meaning-preserved substitutions across multiple languages.

\textbf{LLMs.} The results of Davinci-002 and Davinci-002$+$ from OpenAI are from the paper \cite{aumiller-gertz-2022-unihd}. Davinci-002 is a zero-shot method with a prompt asking for simplified synonyms given a particular context. Davinci-002$+$ is an ensemble over six different prompts/configurations with average rank aggregation. 

We conduct the experiment using a 3-shot prompting method, with the prompts shown in Figure \ref{fig:LS_prompt}. The examples used in the prompt are sourced from the validation set of TSAR. 

\begin{figure}
\centering
\begin{boxedminipage}{\columnwidth}
\footnotesize
\#\#\#Instruction\#\#\#\\
Given a sentence containing a complex word, you should return an ordered list of "simpler" valid substitutes for the complex word in its original context. Valid substitutes are words that are simpler than the complex word and preserve the meaning of the sentence when used as a replacement. The list of simpler words (up to a maximum of 10) should be ordered by your confidence in the prediction (best predictions first). The ordered list must not contain ties.\\
$[(examples)]$\\
\#\#\#Task\#\#\#\\
Context:\\ 
Complex Word:\\
Valid substitutes:
\end{boxedminipage}
\caption{Prompt template for LexS.}
\label{fig:LS_prompt}
\end{figure}

\textbf{(2) Experimental Setting}

\textbf{Dataset:} The TSAR-2022 shared task \cite{saggion2023findings} includes instances in three languages—English, Spanish, and Portuguese—each containing a complex word and its corresponding list of simplified substitutes. We have selected the test set from this dataset as the evaluation corpus for this study, with its statistical information presented in Table \ref{tab:Lexical_statistics}.
\begin{table}[]
\centering
\begin{tabular}{c|c|ccc}
\hline
\multirow{2}{*}{Dataset} & \multirow{2}{*}{Instances} & \multicolumn{3}{c}{Substitution per target} \\ \cline{3-5} 
 &  & Min & Max & Avg \\ \hline
English & 386 & 2 & 22 & 10.55 \\
Spanish & 381 & 2 & 19 & 10.28 \\
Portuguese & 386 & 1 & 16 & 8.10 \\ \hline
\end{tabular}
\caption{Statistics from the LexS dataset we used.}
 \label{tab:Lexical_statistics} 
\end{table}

\textbf{Metrics:} We use the same evaluation metrics as the TSAR-2022 shared task to assess the performance of LexS methods across three languages: ACC@1, Accuracy@n@top1, MAP@k, and Potential@k, $n \in {1,2,3}$ and $ k \in {3,5,10}$. 

(1) Accuracy@n@top1 evaluates whether the most frequently occurring simplified word in the label is included among the top-k predicted candidates. 

(2) Potential@k is defined as the presence of at least one word from the human-annotated simplified word list within the top-k alternatives provided by the model. 

(3) MAP@k additionally takes into account the rank of relevant alternatives among the top-k generated candidates. 

(4) ACC@1 is equivalent to both Potential@1 and MAP@1. 

\textbf{(3) Results} 

\begin{table*}[]
\centering
\begin{tabular}{c|c|c|ccc|ccc|ccc}
\hline
\multirow{2}{*}{Lang} & \multirow{2}{*}{Method} & \multirow{2}{*}{ACC@1} & \multicolumn{3}{c|}{Acc@n@Top1} & \multicolumn{3}{c|}{MAP@k} & \multicolumn{3}{c}{Potential@k} \\ \cline{4-12} 
 &  &  & n=1 & n=2 & n=3 & k=3 & k=5 & k=10 & k=3 & k=5 & k=10 \\ \hline
\multirow{8}{*}{English} & LSBERT & 0.5978 & 0.3029 & 0.4450 & 0.5308 & 0.4079 & 0.2957 & 0.1755 & 0.8230 & 0.8766 & 0.9463 \\
 & MANTIS & 0.6568 & 0.3029 & 0.4450 & 0.5388 & 0.4730 & 0.3599 & 0.2193 & 0.8766 & 0.9463 & 0.9785 \\
 & LSPG & 0.8176 & 0.4557 & 0.6166 & 0.6890 & 0.5881 & 0.4632 & 0.2994 & 0.9624 & 0.9839 & \textbf{0.9973} \\
 \cdashline{2-12}
 & Davinci-002 & 0.7721 & 0.4262 & 0.5335 & 0.5710 & 0.5096 & 0.3653 & 0.2092 & 0.8900 & 0.9302 & 0.9436 \\
 & Davinci-002+ & 0.8096 & 0.4289 & 0.6112 & 0.6863 & 0.5834 & 0.4491 & 0.2812 & 0.9624 & 0.9812 & 0.9946 \\
  \cdashline{2-12}
  & Gemma2(2B) & 0.7479 & 0.3941 & 0.5522 & 0.6461 & 0.5522 & 0.4099 & 0.2415 & 0.9142 & 0.9597 & 0.9785 \\
 & Llama3.1(70B) & 0.8284 & 0.5067 & 0.6541 & 0.7265 & 0.6394 & 0.5019 & 0.3114 & 0.9544 & 0.9758 & 0.9839 \\
 & GPT-4o & \textbf{0.9195} & \textbf{0.5630} & \textbf{0.6997} & \textbf{0.7560} & \textbf{0.7193} & \textbf{0.5535} & \textbf{0.3515} & \textbf{0.9865} & \textbf{0.9946} & 0.9946 \\ \hline
\multirow{8}{*}{Spanish} & LSBERT & 0.2880 & 0.0951 & 0.1440 & 0.1820 & 0.1868 & 0.1346 & 0.0795 & 0.4945 & 0.6114 & 0.7472 \\
 & PresiUniv & 0.3695 & 0.2038 & 0.2771 & 0.3288 & 0.2145 & 0.1499 & 0.0832 & 0.5842 & 0.6467 & 0.7255 \\
 & LSPG & 0.7119 & 0.3722 & 0.5123 & 0.5951 & 0.4983 & 0.3840 & 0.2275 & 0.8831 & 0.9184 & 0.9402 \\
 \cdashline{2-12}
 & Davinci-002 & 0.5706 & 0.3070 & 0.3967 & 0.4510 & 0.3526 & 0.2449 & 0.1376 & 0.6902 & 0.7146 & 0.7445 \\
 & Davinci-002+ & 0.6521 & 0.3505 & 0.5108 & 0.5788 & 0.4281 & 0.3239 & 0.1967 & 0.8206 & 0.8885 & 0.9402 \\
  \cdashline{2-12}
  & Gemma2(2B) & 0.4918 & 0.2500 & 0.3586 & 0.4184 & 0.3164 & 0.2341 & 0.1340 & 0.7445 & 0.8206 & 0.8559 \\
 & Llama3.1(70B) & 0.7282 & 0.3885 & 0.4972 & 0.5461 & 0.4847 & 0.3618 & 0.2118 & 0.8913 & 0.9239 & 0.9429 \\
 & GPT-4o& \textbf{0.8396} & \textbf{0.5108} & \textbf{0.6711} & \textbf{0.7527} & \textbf{0.6180} & \textbf{0.4686} & \textbf{0.2826} & \textbf{0.9592} & \textbf{0.9755} & \textbf{0.9864} \\ \hline
\multirow{8}{*}{Portuguese} & LSBERT & 0.3262 & 0.1577 & 0.2326 & 0.286 & 0.1904 & 0.1313 & 0.0775 & 0.4946 & 0.5802 & 0.6737 \\
 & GMU-WLV & 0.4812 & 0.2540 & 0.3716 & 0.3957 & 0.2816 & 0.1966 & 0.1153 & 0.6871 & 0.7566 & 0.8395 \\
 & LSPG & 0.7433 & 0.4598 & 0.5989 & 0.6524 & 0.5023 & 0.3739 & 0.2250 & 0.9197 & 0.9491 & 0.9625 \\
 \cdashline{2-12}
 & Davinci-002 & 0.6363 & 0.3716 & 0.4615 & 0.5160 & 0.4105 & 0.2889 & 0.1615 & 0.7860 & 0.8181 & 0.8422 \\
 & Davinci-002+ & 0.7700 & 0.4358 & 0.5347 & 0.6229 & 0.5014 & 0.3620 & 0.2167 & 0.9171 & 0.9491 & 0.9786 \\
 \cdashline{2-12}
  & Gemma2(2B) & 0.4786 & 0.2486 & 0.3636 & 0.4438 & 0.3177 & 0.2281 & 0.1318 & 0.7780 & 0.8262 & 0.8770 \\
 & Llama3.1(70B) & 0.7754 & 0.4278 & 0.5561 & 0.6310 & 0.5072 & 0.3640 & 0.2132 & 0.9411 & 0.9679 & 0.9759 \\
 & GPT-4o & \textbf{0.8850} & \textbf{0.5347} & \textbf{0.6684} & \textbf{0.7326} & \textbf{0.5886} & \textbf{0.4448} & \textbf{0.2696} & \textbf{0.9786} & \textbf{0.9893} & \textbf{0.9973} \\ \hline
\end{tabular}
\caption{Evaluation Results of LexS task on English, Spanish, and Portuguese. }
 \label{tab:lexical_result} 
\end{table*}

The results of all methods are displayed in Table \ref{tab:lexical_result}. The results highlight a clear performance gap between traditional non-LLM methods like LSBERT and LSPG and large-scale LLMs like Llama3.1 and GPT-4o. The performance of lightweight LLM (Gemma2) is poorer and lower than non-LLM method (LSPG).

In English, GPT-4o achieves an impressive ACC@1 of 0.9195, outperforming all other methods by a considerable margin, with Llama3.1 and GPT3+ following as the next best-performing methods. Similar trends are observed in Spanish and Portuguese, where GPT-4o again secures the highest ACC@1 scores of 0.8396 and 0.885, respectively. These results highlight the strong capability of GPT-4o in identifying the most appropriate simplified lexical substitutes while maintaining high relevance and potential for simplification tasks. Notably, the performance of non-LLM-based methods declines substantially in Spanish and Portuguese. In contrast, LLM-based method demonstrate more consistent cross-linguistic performance, highlighting their advantages in handling multilingual TS tasks.

\subsection{Sentence Simplification (SenS)}

\textbf{(1) Baselines} 

Most SenS work treats the task as a monolingual machine translation problem, training models on datasets containing complex-simple sentence pairs by sequence-to-sequence modeling \cite{xu2016optimizing,LuQLYZ21}. Some previous work has investigated the simplification capabilities of select LLMs in order to benchmark performance against dedicated approaches \cite{Feng-abs-2302-11957,kew2023bless}.

\textbf{MUSS.} MUSS \cite{martin-etal-2022-muss} is chosen as our primary baseline due to its demonstrated state-of-the-art performance. It fine-tunes a BART-large model \cite{lewis-etal-2020-bart} using ACCESS control tokens extracted from labeled TS datasets and/or mined paraphrases, enabling the training of both supervised (MUSS-wiki-mined) and unsupervised (MUSS-mined) TS systems. 

\textbf{LLMs.} We use OpenAI's Davinci-002 and Davinci-003 LLMs from \cite{kew2023bless} as baselines and adopt prompt2 from \cite{kew2023bless} as our prompt, illustrated in Figure \ref{fig:SS_prompt}. For all LLM settings, three complex-source pairs from the validation sets are randomly sampled as few-shot examples.

\begin{figure}
\centering
\begin{boxedminipage}{\columnwidth}
\footnotesize

Please rewrite the following complex sentence in order to make it easier to understand by non-native speakers of English. You can do so by replacing complex words with simpler synonyms (i.e.paraphrasing), deleting unimportant information (i.e.compression), and/or splitting a long complex sentence into several simpler ones. The final simplified sentence needs to be grammatical, fluent, and retain the main ideas of its original counterpart without altering its meaning.\\
$[(examples)]$\\
Complex:\\ 
\end{boxedminipage}
\caption{Prompt template for SenS.}
\label{fig:SS_prompt}
\end{figure}

\textbf{(2) Experimental Setting}

\begin{table*}[]
\centering
\begin{tabular}{c|c|c|c|c|c}
\hline
Dataset & Method & SARI(↑) & BERTScore(↑) & FKGL(↓) & LENS(↑) \\ \hline
\multirow{8}{*}{ASSET} & Gold References & 45.27 & 78.89 & 6.53 & 65.58 \\
\cdashline{2-6}
 & MUSS-mined & 42.29 & 79.86& 8.18 & 61.36 \\
 & MUSS-wiki-mined & 44.90 & 77.71 & \textbf{5.29} & 69.23 \\
 \cdashline{2-6}
 & Davinci-002 & 42.84 & 85.91 & 7.77 & 67.09 \\
 & Davinci-003 & 46.60 & 79.66 & 7.74 & 67.39 \\
 \cdashline{2-6}
 & Gemma2(2B) & 46.20 & 82.90 & 6.53 & 72.44 \\
 & Llama3.1(70B) & 47.27 & 79.80 & 5.62 & \textbf{74.64} \\
 & GPT-4o & \textbf{49.20} & \textbf{86.38} & 7.20 & 72.28 \\ \hline
\multirow{8}{*}{MED-EASI} & Gold References & 100.00 & 100.00 & 9.59 & 65.89 \\
\cdashline{2-6}
 & MUSS-mined & 35.15 & 42.55 & 9.29 & 52.48 \\
 & MUSS-wiki-mined & 35.12 & 43.07 & 8.04 & 59.12 \\
 \cdashline{2-6}
 & Davinci-002 & 36.34 & \textbf{43.67} & 43.67 & 57.71 \\
 & Davinci-003 & 39.81 & 40.83 & 7.12 & 46.80 \\
 \cdashline{2-6}
 & Gemma2(2B) & 38.91 & 40.50 & 7.67 & 66.57 \\
 & Llama3.1(70B) & 39.55 & 39.41 & \textbf{6.86} & \textbf{69.12} \\
 & GPT-4o & \textbf{40.81} & 39.89 & 7.73 & 65.92 \\ \hline
\end{tabular}
\caption{ Evaluation results of SenS on ASSET and MED-EASI datasets.}
 \label{tab:Sentence_result} 
\end{table*}

\textbf{Dataset:} Our experiments are conducted on two datasets, encompassing two different domains:

ASSET is the most widely used SenS dataset, containing 2,359 sentences from English Wikipedia, each paired with 10 simplified references. For evaluation, we use the official test split consisting of 359 sentences. The simplified references were generated by crowdworkers who were guided to apply edit operations such as replacement, splitting, and deletion.

MED-EASI is a specialized dataset designed for the simplification of concise medical texts. It contains 1,979 pairs of complex (expert) and simplified (layman) texts. The dataset was collaboratively annotated by experts, non-experts, and AI systems, supporting four types of text transformations: elaboration, substitution, deletion, and insertion, to achieve controlled TS. We used its test set, which includes 300 test samples, for evaluation.

\textbf{Metrics:} To evaluate the performance of the models, we employed a series of automatic evaluation metrics. 

SARI \cite{xu2016optimizing} is commonly used to assess the quality of simplification, which calculates the F1 score for the addition, rephrase, and deletion of n-grams relative to the output and reference sentences. 

BERTScore \cite{papineni2002bleu} is utilized to measure the similarity between the output and the reference simplified sentences. 

FKGL (Flesch-Kincaid Grade Level) \cite{kincaid1975derivation} is employed to compute the readability of the text, which is a weighted score based on sentence length and syllable count. A lower FKGL score indicates simpler output. 

One recently proposed LENS \cite{maddela2023lens} metric is used, which considers both the semantic similarity of the output to the source and reference sentences, as well as the degree of simplification achieved.

\textbf{(3) Results}

The results of LLMs compared to the previous state-of-the-art methods are presented in Table \ref{tab:Sentence_result}. The results highlight a clear distinction between the capabilities of non-LLM approaches, such as MUSS-mined and MUSS-wiki-mined, and LLM-based methods. Non-LLM approaches, despite their historical significance and optimized pre-training for specific tasks, show limitations in both generalization and adaptability across domains. While MUSS-wiki-mined achieves competitive FKGL scores in the Wikipedia domain, it fails to maintain a strong balance between simplification, fluency, and readability, as evidenced by lower SARI and BERTScore scores. These shortcomings are even more pronounced in the specialized medical domain, where the demand for domain-specific knowledge and contextual understanding exceeds the capabilities of smaller, task-specific models.

GPT-4o and Llama3.1 including lightweight LLM (Gemma2) achieve superior SARI scores, indicating their capacity to simplify sentences while preserving meaning and key information. Their high LENS scores suggest these models successfully integrate multiple aspects of simplification, offering outputs that are both accessible and natural. 

\subsection{Syntactic Simplification (SynS)}

\textbf{(1) Baselines} 
Prior to LLMs, the work of SynS can be divided into two categories. The first one is modeled as a sequence-to-sequence task where systems are trained on parallel corpora synthesized from knowledge graphs \cite{narayan-etal-2017-split}, mined from Wikipedia \cite{botha2018learning} and crowd-sourced \cite{gao2021abcd}. The second one relies on a larger set of expert-crafted lexical rules \cite{niklaus2023discourse,yao2024semantic}.

\textbf{DisSim.} DisSim \cite{niklaus2023discourse} applies a recursive transformation to sentences using a set of 35 hand-crafted syntactic and lexical rules based on phrase structure.

\textbf{ABCD.} ABCD \cite{gao2021abcd} represents sentences as graphs, with edges capturing dependency and neighboring relations, and trains a neural network to predict actions on these edges.

\textbf{DSS.} DSS \cite{sulem2018simple} utilizes UCCA \cite{abend2013ucca} for semantic representation, splits the UCCA graph into subgraphs based on parallel and elaborator scenes, and generates text from the subgraphs using a neural model.

\textbf{AMRS3.} AMRS3 \cite{yao2024semantic} leverages AMR \cite{banarescu2013abstract} as the semantic representation, decomposing the AMR graph of a complex sentence into a series of subgraphs. These subgraphs then guide the generation of simpler sentences, which are combined to produce the final output.

\textbf{LLMs.} We choose these LLMs (GPT-3.5 and Llama-3-8B) in \cite{yao2024semantic} for comparison,  where the version of GPT-3.5 is "gpt-3.5-turbo-0125". We select zero-shot strategy, and the specific prompt used \cite{yao2024semantic} is illustrated in Figure \ref{fig:Syntactic_prompt}.

\begin{figure}
\centering
\begin{boxedminipage}{\columnwidth}
\footnotesize
You are a helpful assistant that simplifies syntactic structures. Rewrite the following paragraph using simple sentence structures and no clauses or conjunctions: ${[sentence]}$ 
\end{boxedminipage}
\caption{Prompt template for SynS.}
\label{fig:Syntactic_prompt}
\end{figure}

\textbf{(2) Experimental Setting}

\textbf{Dataset:} Our experiments are conducted on one common SynS dataset (WEBSPLIT) \cite{narayan-etal-2017-split}. This dataset contains 1,445,159 complex-simple pairs with partitions, extracted from the final version of the WebNLG corpus and encompassing 15 DBPedia categories. We utilized its test set for evaluation purposes.

\textbf{Metrics:} We select two commonly used metrics for evaluation (BERTScore and L2SCA). BERTScore is used to assess whether the simplified sentence sufficiently retains the meaning of the original sentence. We employ BERTScore Recall, computed using DeBERTa-NLI1 \cite{he2020deberta}, to evaluate the results. 

L2SCA \cite{lu2010automatic} is utilized to measure syntactic complexity, providing automated measurements of 14 syntactic complexity indices across five categories. Following the approach of \cite{yao2024semantic}, we select one index from each of the five categories: MLT (mean length of T-unit), C/S (sentence complexity ratio), C/T (T-unit complexity ratio), T/S (sentence coordination ratio), and CN/T (complex nominals per T-unit). A T-unit refers to a syntactic unit consisting of one independent main clause (a clause that can stand alone as a sentence) and all its dependent subordinate clauses (clauses that cannot stand alone as sentences).

\textbf{(3) Results}

\begin{table*}[]
\centering
\begin{tabular}{l|cc|ccccc}
\hline
\multirow{2}{*}{Method} & \multicolumn{2}{c|}{BERTScore(↑)} & \multicolumn{5}{c}{L2SCA(↓)} \\ \cline{2-8} 
 & Mean & Median & MLT & C/S & C/T & T/S & CN/T \\ \hline
Exact Copy & 1.00 & 1.00 & 16.57 & 1.64 & 1.50 & 1.10 & 1.72 \\
\cdashline{1-8}
ABCD & 0.90 & \textbf{0.91} & 9.53 & 1.00 & 1.10 & 0.91 & 0.94 \\
DisSim & 0.87 & 0.87 & 8.54 & 1.05 & 1.05 & 0.99 & 0.67 \\
DSS & 0.74 & 0.74 & 10.69 & 0.97 & 1.19 & \textbf{0.81} & 1.05 \\
AMRS3 & 0.81 & 0.81 & 8.92 & 1.00 & 1.02 & 0.99 & 0.68 \\
\cdashline{1-8}
GPT-3.5 & \textbf{0.90} & 0.90 & 7.79 & 1.02 & 1.02 & 1.00 & 0.52 \\
Llama-3(8B) & 0.84 & 0.85 & \textbf{6.69} & 1.01 & 1.01 & 1.00 & \textbf{0.38} \\
\cdashline{1-8}
Gemma2(2B) & 0.83 & 0.85 & 7.28 & 1.03 & 1.03 & 1.00 & 0.48 \\
Llama3.1(70B) & 0.85 & 0.85 & 7.49 & \textbf{1.00} & 1.01 & 0.99 & 0.60 \\
GPT-4o & 0.86 & 0.87 & 6.82 & \textbf{1.00} & \textbf{1.00} & 0.99 & 0.46 \\ \hline
\end{tabular}
\caption{Evaluation results of SynS on WEBSPLIT dataset.}
 \label{tab:Syntactic_result} 
\end{table*}

The results of LLMs compared to the previous state-of-the-art methods are presented in Table \ref{tab:Syntactic_result}. In terms of meaning preservation on the WEBSPLIT dataset, the simplifications generated by LLMs are comparable to or better than those produced by traditional methods. Among them, ABCD and GPT-3.5 achieve the highest scores. However, ABCD requires additional specialized training, whereas LLMs can achieve excellent performance with only one simple prompt.

LLMs in zero-shot settings, demonstrate strong performance in SynS without a substantial loss in meaning. Llama-3(8B), for instance, produces the simplest outputs with the lowest mean length of T-unit (MLT = 6.69) and complex nominals per T-unit (CN/T = 0.38), although it shows a slight dip in semantic retention (BERTScore around 0.84–0.85). Overall, the results underscore the trade-offs between simplification and semantic preservation, highlighting that modern LLMs can effectively balance these aspects compared to traditional rule-based and graph-based methods.

\subsection{Document Simplification (DocS)}

\textbf{(1) Baselines}

DocS is a very challenging task because text generation for long sequences was very difficult before the advent of LLMs. One of the existing DocS uses sentence or paragraph simplification \cite{laban-etal-2021-keep} and the other is to perform specific sentence transformations \cite{zhang-etal-2022-predicting,cripwell2023document}.

\textbf{Keep it Simple (KIS).}  A multi-paragraph level unsupervised method for TS \cite{laban-etal-2021-keep}.

\textbf{BART-SWIPE.} A model fine-tuned on a cleaned version of SWIPE, a large-scale document-level simplification dataset based on Wikipedia, constructs pairs of documents by gathering pages from both English and Simple English Wikipedia \cite{Laban2023SWiPEAD}.

\textbf{PG.} A plan-guided (PG) system is implemented where a planner predicts an operation for each sentence and provides it as a control token to a sentence-level  simplification model \cite{cripwell2023document}.

\textbf{(3) LLMs.} We choose two results of GPT-3.5 and GPT-4\cite{fang-etal-2025-collaborative}, where the versions of GPT-3.5 and GPT-4 are "gpt-3.5-turbo-0125" and "gpt-4-0125-preview".

We selected Prompt Template 2 from \cite{fang-etal-2025-collaborative} as our prompt, this prompt follows a zero-shot prompting strategy. and the specific prompt used is illustrated in Figure \ref{fig:DS_prompt}.

\begin{figure}
\centering
\begin{boxedminipage}{\columnwidth}
\footnotesize
As a text simplification writer, your task is to simplify the given text content:
restate the original text in simpler and easier to understand language without changing its meaning as much as possible.\\
You can change paragraph or sentence structure, remove some redundant information, and replace complex and uncommon expressions with simple and common ones.\\
It should be noted that the task of text simplification is completely different from the task of text summarization, so you need to provide a simplified parallel version based on the original text, rather than just providing a brief summary.\\
Raw text:$[Raw\ text]$\\
Simplified text:
\end{boxedminipage}
\caption{Prompt template for document simplification.}
\label{fig:DS_prompt}
\end{figure}

\textbf{(2) Experimental Setting}

\textbf{Dataset.} We utilize one commonly used dataset, Newsela \cite{xu-etal-2015-problems}. This corpus comprises thousands of news articles that have been professionally leveled to different reading complexities. Each document includes four reference simplified versions. Due to the costs associated with running LLMs via APIs, we adopted the same processing method as \cite{fang-etal-2025-collaborative}, randomly sampling 200 documents from the original dataset for our experiments.

\textbf{Metrics.} Based on factors such as simplicity, completeness, and fluency, we select four automatic evaluation metrics. SARI and FKGL in SenS are also chosen.

D-SARI \cite{sun-etal-2021-document} is an improved metric based on SARI, which penalizes the three components of SARI and is particularly suitable for document-level simplification evaluation. 

BARTScore \cite{Yuan2021BARTScoreEG} is employed to evaluate the meaning preservation and fluency of the generated text. 

\textbf{(3) Results}

\begin{table*}[]
\centering
\begin{tabular}{c|cccc}
\hline
Method & SARI(↑) & D-SARI(↑) & BARTScore(↓) & FKGL(↓) \\ \hline
KIS & 33.26 & 26.58 & -2.92 & 9.32 \\
BART-SWIPE & 30.23 & 23.78 & -3.16 & 8.58 \\
PG & 36.52 & 27.31 & -3.18 & 7.85 \\
\cdashline{1-5}
GPT-3.5 & 32.38 & 22.71 & -2.45 & 7.81 \\
GPT-4 & 33.61  & 22.67 & -2.78 & 7.58\\
\cdashline{1-5}
Gemma2(2B) & 29.55 & 22.16 & -3.90 & 10.24 \\
Llama3.1(70B) & 36.70 & 25.78 & \textbf{-2.28} & 8.30 \\
GPT-4o & \textbf{41.96} & \textbf{29.60} & -2.54 & \textbf{5.46} \\ \hline
\end{tabular}
\caption{Evaluation results of Document simplification on Newsela dataset. }
 \label{tab:Document_result} 
\end{table*}

The results of LLMs compared to the previous state-of-the-art methods are presented in Table \ref{tab:Document_result}. GPT-4o stands out with the highest SARI (41.96) and D-SARI (29.60) scores, alongside the best FKGL score (5.46), indicating that it not only simplifies text effectively but also produces output that is easier to read. Its competitive BARTScore (–2.54) further confirms its ability to maintain meaning and fluency despite significant simplification. These results suggest that GPT-4o is particularly adept at managing the inherent challenges of DocS, especially the generation of coherent long-form simplified text, marking a substantial improvement over previous techniques that relied on sentence or paragraph-level transformations.

In comparison, Llama3.1 demonstrates robust performance as an open-source alternative, with a solid SARI (36.70) and D-SARI (25.78) score, as well as a comparable BARTScore (–2.28), though its FKGL (8.30) indicates a somewhat higher reading level than GPT-4o’s output. Meanwhile, the lightweight Gemma2-2B model lags behind, recording the lowest SARI (29.55) and D-SARI (22.16) scores and a higher BARTScore (–3.90) alongside the most complex output per FKGL (10.24).

\subsection{Human Evaluation}

In the above automatic evaluations, they compare a simplification text against its human-created reference text. TS metrics must give the maximum score to the reference text. However, these metrics overlook the potential for a candidate text to exceed the reference text in terms of quality. In particular, recent advancements in LLMs have highlighted this issue \cite{noh2024beyond}, as LLM-generated texts often exceed the quality of human-written texts. Therefore, we design a simple manual evaluation experiment. Given the original text, a human is asked to determine which is better, the manually labeled results or the results generated by GPT-4o.

For each test set, we randomly select 100 instances from English(LexS), ASSET(SenS), WEBSPLIT(SynS) and Newsela(DocS). Given the original text, a human is asked to determine which is better from multiple aspects, the manually labeled results or the results generated by GPT-4o. We recruited three graduate students who are non-native speakers to evaluate separately, and then we calculated the average. Here's a shorter explanation of each evaluation aspect:

\textbf{Accuracy for LexS:} Evaluates whether the simplified words generated in LexS meet two core criteria: firstly, the simplified word should be more comprehensible than the complex word; secondly, substituting the original word should not alter the meaning or the informational integrity of the text.

\textbf{Comprehensiveness (Comp) for LexS:} In the context of LexS, a complex word may have multiple potential simplified alternatives. This metric aims to evaluate the extent to which a system's generated simplification results cover all possible reasonable substitution options.

\textbf{Coherence (Coher) for SynS and DocS:} Evaluates the logical flow and organization of the simplified text, ensuring smooth transitions and interconnectedness between sentences and paragraphs.

\textbf{Faithfulness (Faith) for SenS, Syns and DocS:} Measures how well the simplified text preserves the core meaning, key information, and intended message of the original text without distorting or misrepresenting them.

\textbf{Simplicity (Simp) for all tasks:} Assesses the level of complexity and difficulty in the simplified text, aiming to make the content more accessible through plain language, shorter sentences, and simpler vocabulary.

\textbf{Overall Evaluation for all tasks:} From all the metrics considered for each task, the most suitable answer is selected, which, from the annotator's perspective, is deemed to best meet the simplification requirements.

\begin{table}[]
\centering
\begin{tabular}{c|c|c|c|c|c|c}
\hline
\multirow{2}{*}{LexS} & Accu & Comp & Simp & Overall \\
 & 58.3 & 55.0 & 51.3 & 55.3  \\ \hline
 \multirow{2}{*}{SenS} & Coher & Faith & Simp & Overall \\
 & 55.3 & 58.7 & 44.3 & 59.3  \\ \hline
\multirow{2}{*}{SynS} & - & Faith & Simp & Overall \\
 & & 79.3 & 86.3 & 84.7  \\ \hline
\multirow{2}{*}{DocS} & Coher & Faith & Simp & Overall \\
 & 59.3 & 52.3 & 65.7 & 63.7  \\ \hline
\end{tabular}
\caption{The results of human judgment on the quality of GPT-4o and human annotations. Each number represents the number of instances in which the GPT-4o results are better than the human annotations out of 100 instances.}
 \label{tab:Human_Evaluation_Result} 
\end{table}

The human evaluation results reveal that GPT-4o frequently outperforms human annotations across all tasks. In SynS, GPT-4o excels significantly, achieving high scores in faithfulness (79.3\%), simplicity (86.3\%), and overall quality (84.7\%), suggesting its superiority in restructuring sentences while preserving meaning and clarity. For LexS, GPT-4o marginally surpasses humans in accuracy (58.3\%) and comprehensiveness (55\%) but lags slightly in simplicity (51.3\%), indicating room for improvement in balancing lexical substitution with accessibility. DocS shows GPT-4o’s strength in simplicity (65.7\%) and coherence (59.3\%), though faithfulness (52.3\%) remains a weaker point, hinting at occasional trade-offs between clarity and fidelity. Conversely, in SenS, while GPT-4o maintains better faithfulness (58.7\%) and coherence (55.3\%), humans retain an edge in simplicity (44.3\%), underscoring their nuanced understanding of audience-specific readability.

These findings challenge the assumption that human references are inherently superior, particularly in structural tasks (SynS, DocS) where GPT-4o’s systematic capabilities shine. The results advocate for revising automatic evaluation metrics to account for scenarios where LLM-generated simplifications exceed human quality, while highlighting the continued value of human judgment in fine-grained linguistic adjustments.

\subsection{Discussion}

\textbf{(1) LLMs Achieves Superior Performance:}
The experimental results show that the commercial GPT‑4o outperforms all other systems in all tasks using automatic metrics. Notably, GPT‑4o even surpasses manually labeled (human-annotated) results via human evaluations, underscoring its exceptional ability to produce text that is both clear and structurally simplified. Llama‑3.1 exhibits strong performance that is very competitive with the best commercial models. As the performance of LLMs continues to improve, their ability in TS will also only get better, something that traditional non-large language models cannot compare to. 

\textbf{(2) The lightweight Gemma‑2 Excels in SenS and Syns:}
Lightweight Gemma‑2 model shows clear improvements over traditional, non‑LLM approaches—but only in the areas of SenS and SynS. While Gemma‑2 successfully simplifies sentences and reduces syntactic complexity better than older methods, its performance in tasks that require broader language understanding, such as lexical and document simplification, does not match that of its larger counterparts. This finding highlights that while smaller LLMs can be effective for sentence-level transformation, their limited capacity may restrict their overall utility for more complex simplification tasks.

\textbf{(3) Limitations of Traditional Evaluation Metrics:}

The results also reveal that existing evaluation metrics, which are largely based on manual annotations, fall short when assessing the capabilities of modern LLMs. As LLMs like GPT‑4o can produce outputs that exceed the quality of human-created references in terms of simplicity, fluency, and structural clarity, traditional metrics struggle to capture these advancements adequately. This gap indicates a pressing need to develop new evaluation methodologies that better align with the high performance and nuanced output of contemporary LLMs, ensuring that their strengths are accurately measured and appreciated.

\section{Future Directions}

It can be seen that LLMs have already achieved outstanding performance in existing text simplification tasks. However, there are still some very challenging tasks in text simplification that have not received enough attention. Below are some possible research directions.

\subsection{Multi-Level Text Simplification}

For readers with different reading levels, many literary works have multiple simplified versions of varying levels. The aim of multi-level TS \cite{spring2021exploring} is to generates multiple simplified versions for the needs of users with different reading abilities. Simplification at different levels is crucial for tailoring content to diverse audiences—ranging from children and non-native speakers to individuals with cognitive or learning disabilities. For instance, an academic article might need a mild simplification for college students but a much deeper rewrite for younger readers or people with low literacy skills. This differentiation can enhance understanding, improve educational outcomes, and promote inclusivity. 

However, multi-level TS also presents significant challenges. One major issue is ensuring consistency and coherence across levels. For example, making sure that lexical simplifications do not conflict with higher-level structural changes. Integrating modifications at the word, sentence, and discourse levels requires sophisticated coordination and often iterative processing, which can be computationally demanding. Furthermore, as models work at different granularities, they must avoid oversimplification that could result in loss of critical context or even introduce factual inaccuracies. Finally, designing effective evaluation metrics that accurately capture improvements across these multiple levels remains an open research problem, underscoring the complexity and interdisciplinary nature of this future task.

\subsection{Personalized Text Simplification}

Personalized TS \cite{bingel2018lexi} aims to tailor the simplification process according to individual readers’ needs, preferences, and reading abilities. This customization is crucial because people differ significantly in language proficiency, cognitive processing, and domain-specific knowledge. For instance, while a non-native speaker might benefit from simpler vocabulary substitutions, a child may need both simpler vocabulary and restructured sentence patterns for better comprehension. By adapting to the user’s profile, personalized TS can enhance accessibility and engagement, ensuring that the simplified text not only retains the core meaning of the original document but also resonates with the reader’s personal context and learning requirements.

However, developing effective personalized TS systems comes with several challenges. One of the primary difficulties is accurately modeling individual reader profiles, which may involve diverse factors such as age, educational background, language proficiency, and even specific reading interests. This requires collecting and processing personalized data while addressing privacy concerns and ensuring ethical data usage. Additionally, integrating personalization into the simplification process demands sophisticated algorithms that can dynamically adjust multiple levels of text complexity (lexical, syntactic, and discourse) in a coordinated manner. Finally, evaluating personalized simplification poses its own challenges, as standard metrics may not fully capture the user-specific improvements, calling for the development of new evaluation frameworks that incorporate human-in-the-loop assessments and individualized feedback.

\subsection{Bridging the Gap: Empowering Lightweight LLMs with Large-Scale Simplification Capabilities}

Achieving the simplifying power of lightweight LLMs is critical for democratizing TS. Lightweight LLMs promise faster inference, lower deployment costs, and the ability to operate in resource-constrained environments (e.g., mobile devices or edge computing) while ensuring data privacy. If it can be endowed with the nuanced simplification capabilities of LLMs, they could enable high-quality, personalized, and real-time simplification services for diverse user groups—from non-native speakers to individuals with learning disabilities—without the heavy computational overhead of larger models. Moreover, such advances would facilitate on-device applications and broaden access to simplified content globally, making digital information more universally comprehensible.

Key challenges in equipping lightweight LLMs with the power of LLMs include maintaining semantic fidelity and managing complex contextual dependencies while operating within a reduced parameter budget. Knowledge distillation, compression techniques, and efficient fine-tuning are promising approaches, yet they must balance between preserving the subtle linguistic nuances (such as idiomatic expressions and discourse coherence) and preventing oversimplification that might lead to loss of critical information. The trade-offs between model size, processing speed, and generalization across diverse text types add additional layers of complexity to this research frontier.

\subsection{Beyond Reference: Evaluating High Quality Simplifications Better than Human References}

Designing evaluation frameworks that surpass human references poses several challenges. First, there is the difficulty of establishing clear, objective criteria that capture the multifaceted nature of TS, including readability, semantic fidelity, and stylistic appropriateness. Standard automatic metrics such as SARI, BLEU, and FKGL have limitations and may not reflect human judgments accurately when outputs are of very high quality. Secondly, gathering comprehensive human evaluations is both resource-intensive and subjective, making it hard to create a universally accepted gold standard. Moreover, as simplification systems evolve, the evaluation metrics must adapt to assess nuanced improvements without being biased toward certain linguistic styles or simplification strategies. Balancing these factors while ensuring scalability and consistency across various domains and languages remains an ongoing and complex research challenge.

\section{Conclusions}

This paper provides a comprehensive exploration of LLMs in the field of TS, evaluating their performance across four primary tasks: lexical, syntactic, sentence, and document simplification. Our findings demonstrate the superior capabilities of LLMs, especially proprietary models like GPT-4o, in delivering high-quality simplifications that often surpass traditional approaches and even human-annotated references in terms of both precision and contextual retention.

However, this progress highlights critical challenges, such as the need for more robust evaluation metrics and the computational demands of large-scale models. Future research must focus on developing efficient, scalable methods, including knowledge distillation and lightweight models, to democratize access to high-quality simplification technologies. By addressing these challenges and embracing the opportunities presented by LLM advancements, the field of TS is poised to redefine how diverse audiences engage with complex content, setting the stage for further innovations in AI-driven accessibility solutions.

\section*{Acknowledgments}

This research is partially supported by the National Natural Science Foundation of China (62076217 and 62120106008), the National Language Commission (ZDI145-71).

\bibliographystyle{named}
\bibliography{ijcai25}

\end{document}